\documentclass[letterpaper, 10 pt, conference]{ieeeconf}  % Comment this line out if you need a4paper

\IEEEoverridecommandlockouts                              % This command is only needed if 
\usepackage{amsmath,amsfonts, bm}
\usepackage[ruled]{algorithm2e}
\usepackage{array}
\usepackage[caption=false]{subfig}
\usepackage{textcomp}
\usepackage{stfloats}
\usepackage{url}
\usepackage{verbatim}
\usepackage{graphicx}
\usepackage{cite}
\usepackage{xcolor}
\usepackage{booktabs, multirow}       % professional-quality tables
\usepackage{bbding}
\usepackage{multirow}
\usepackage[hidelinks]{hyperref}
\usepackage[utf8]{inputenc}

\iffalse
    \newcommand{\holger}[1]{\noindent}
\else
    \newcommand{\holger}[1]{{\color{blue}[HC: #1]}} % Added by Holger
    
\fi

\newcommand\norm[1]{\left\lVert#1\right\rVert}

\newcounter{daggerfootnote}
\newcommand*{\daggerfootnote}[1]{%
    \setcounter{daggerfootnote}{\value{footnote}}%
    \renewcommand*{\thefootnote}{\fnsymbol{footnote}}%
    \footnote[2]{#1}%
    \setcounter{footnote}{\value{daggerfootnote}}%
    \renewcommand*{\thefootnote}{\arabic{footnote}}%
    }

\title{\LARGE \bf
Label-Efficient 3D Object Detection For Road-Side Units
}

\author{Minh-Quan Dao$^{1}$, Holger Caesar$^{2}$, Julie Stephany Berrio$^{3}$, Mao Shan$^{3}$, \\
        Stewart Worrall$^{3}$,  Vincent Frémont$^{4}$, Ezio Malis$^{1}$   % <-this % stops a space
\thanks{$^{1}$INRIA Centre at Université Côte d'Azur, Sophia Antipolis, France,
        {\tt\small minh-quan.dao@inria.fr}}%
\thanks{$^{2}$Intelligent Vehicles Group, Delft University of Technology, the Netherlands
        }%
\thanks{$^{3}$Australian Centre for Robotics, University of Sydney, Australia,
        }%
\thanks{$^{4}$LS2N, École Centrale de Nantes,
        }%
\thanks{This work has been carried out and funded in the framework of the ANNAPOLIS project managed by the National Agency for Research (ANR-21-CE22-0014)}
}

\SetKwInput{KwInput}{Input}                % Set the Input
\SetKwInput{KwOutput}{Output}              % set the Output

\begin{document}

\maketitle
\thispagestyle{empty}
\pagestyle{empty}

%%%%%%%%%%%%%%%%%%%%%%%%%%%%%%%%%%%%%%%%%%%%%%%%%%%%%%%%%%%%%%%%%%%%%%%%%%%%%%%%
\begin{abstract}
% Navigating intersections poses a significant challenge for  due to the limitations of LiDAR-based perception caused by occlusion.
Occlusion presents a significant challenge for safety-critical applications such as autonomous driving.
Collaborative perception has recently attracted a large research interest thanks to the ability to enhance the perception of autonomous vehicles via deep information fusion with intelligent roadside units (RSU), thus minimizing the impact of occlusion.
While significant advancement has been made, the data-hungry nature of these methods creates a major hurdle for their real-world deployment, particularly due to the need for annotated RSU data. 
Manually annotating the vast amount of RSU data required for training is prohibitively expensive, given the sheer number of intersections and the effort involved in annotating point clouds.
We address this challenge by devising a label-efficient object detection method for RSU based on unsupervised object discovery.
Our paper introduces two new modules: one for object discovery based on a spatial temporal aggregation of point clouds, and another for refinement.
Furthermore, we demonstrate that fine-tuning on a small portion of annotated data allows our object discovery models to narrow the performance gap with, or even surpass, fully supervised models.
Extensive experiments are carried out in simulated and real-world datasets to evaluate our method \protect\daggerfootnote{The code is released in \url{https://gitlab.inria.fr/mdao/label-efficient-rsu}}.
\end{abstract}

%%%%%%%%%%%%%%%%%%%%%%%%%%%%%%%%%%%%%%%%%%%%%%%%%%%%%%%%%%%%%%%%%%%%%%%%%%%%%%%%
\section{INTRODUCTION}
Autonomous vehicles (AVs) have the potential to transform transportation. 
To safely navigate their environment, AVs rely on LiDARs to detect other road users.
While being able to produce accurate and dense range measurements, LiDARs are highly vulnerable to occlusion and sparsity, which result in low-density or absent measurements in certain areas. 
Such unobservable areas can be safety critical as other road users can emerge from these areas, thus risking collisions.
The severity of this issue is quantified by the fact that 80\% of collisions involving AVs in California, USA, occur at intersections where occlusion is the most severe \cite{song2021automated}.

One solution to such a challenge involves enhancing AVs perception through collaboration with intelligent roadside units (RSUs), advanced sensing systems positioned at elevated positions around intersections such that they have a minimally occluded field of view.
Various collaborative perception methods \cite{li2021learning,xu2022v2x} share the common requirement for annotated data to train deep learning models.
The multitude of intersections results in a large amount of data to be annotated.
While human annotation provides the most accurate labels, manual labelling is laborious and costly.
A more scalable method in terms of labelling effort is, therefore, needed.

Recent advances in unsupervised object discovery in outdoor point clouds \cite{you2022learning, wang20224d, najibi2022motion, zhang2023towards, najibi2023unsupervised} present a potential solution to this challenge.
These methods first discover objects with a parametric hand-crafted model made of density-based clustering (e.g., \cite{ester1996density,mcinnes2017hdbscan}) and tightest-box fitting \cite{zhang2017efficient}.
The discovered objects are then utilized as labels for training a deep learning-based detection model according to the self-training process of \cite{xie2020self}.
The main drawback of these methods is their performance gap compared to fully supervised models primarily due to the low precision and recall of the hand-crafted model.

The low recall issue arises from the failure of density-based cluster detection, attributed to low point density and the disjointed nature of point clouds from different parts of the same objects.
Our solution is to use multi-frame multi-scale object discovery is to (1) increase points density by point clouds aggregation using scene flow and, (2) apply clustering algorithms to point clouds at different scales, with smaller point cloud scaling better able to detect larger vehicles (e.g. trucks, busses).

The low precision is the result of poor estimation of the dimension and pose of bounding boxes produced by the tightest-box fitting approach \cite{zhang2017efficient} when clusters form incomplete L-shapes. 
Such incompleteness is caused by objects' partial visibility or the clustering algorithm's missing of objects' points.
We resolve clusters' incompleteness by devising a new refinement method based on the aggregation of points on objects' trajectories.

Furthermore, our unsupervised object discovery is complemented with fine-tuning.
We demonstrate that fine-tuning the model pre-trained using discovered objects and self-training on a small amount of manually labeled data can bridge the performance gap with respect to the fully supervised model.

In summary, our paper makes the following contributions:
\begin{itemize}
    \item We present the first autolabeling framework for RSUs' point clouds to address the challenge of label-efficient object detection models for smart infrastructures.
    \item Our framework introduces two novel modules: (1) object discovery based on spatial and temporal aggregation of point clouds and at multiple scales, and (2) refinement of discovered objects.
    \item We demonstrate that with 100 manually-labeled point clouds, our method achieves performance comparable to models trained fully supervised on 8900 and 1920 manually-labeled point clouds from the synthetic V2X-Sim dataset \cite{li2022v2x} and the real-world dataset A9-Intersection \cite{zimmer2023a9}, respectively, reaching 99\% and 96\% performance, respectively.
\end{itemize}

%%%%%%%%%%%%%%%%%%%%%%%%%%%%%%%%%%%%%%%%%%%%%%%%%%%%%%%%%%%%%%%%%%%%%%%%%%%%%%%%
\section{RELATED WORK}

\subsection{Unsupervised Object Discovery}
Unsupervised Object Discovery is a recent advance of the unsupervised object detection.
Methods of this framework consists of a hand-crafted model for discovering (or localizing) objects in point clouds based on geometrical and statistical cues. 
The discovered objects then play the role of the initial labels set in the training of learning-based detection models following the self-training method of \cite{xie2020self}. 

MODEST~\cite{you2022learning} is the first to demonstrate object discovery in automotive point clouds. 
Its object discovery model is based on the ephemeral score which is a statistic describing the consistency of the neighbourhood of a LiDAR point across different traversals.
A concerntration of points with high ephemeral scores, detected by DBSCAN \cite{ester1996density}, is regarded as a dynamic object, which is then localized by the bounding box fitting algorithm of \cite{zhang2017efficient}.
MODEST has three drawbacks including (1) its need for multiple traversals, (2) its restriction to discover only dynamic objects, and (3) its limited performance compared to fully supervised models.

MI-UP~\cite{najibi2022motion} resolves the multi-traversal requirement by detecting dynamic points using the scene flow.
Furthermore, it improves the final performance by tracking discovered objects through time to obtain a better estimation of their dimension.
While achieving strong performance MI-UP's reliance on dynamic points, like MODEST, confines its scope exclusively to discovering dynamic objects.

OYSTER~\cite{zhang2023towards} overcomes such restrictions on dynamic points by directly applying DBSCAN to non-ground points, although this approach carries the risk of introducing spurious clusters due to static points, primarily from the background.
% Such restriction to dynamic points is overcome by OYSTER~\cite{zhang2023towards} which directly applies DBSCAN to non-ground points at the risk of having spurious clusters caused by static points, predominantly background.
Its solution to spurious clusters is to restrict the discovery range (40 meters for 64-beam LiDARs) so that foreground objects (e.g., cars, pedestrians) have a high density of LiDAR points and complete appearances, thus facilitating density-based clustering.
Furthermore, OYSTER employs tracking to filter objects belonging to short trajectories. 
The detection range is subsequently extended during the self-training stage, thanks to a data augmentation method that drops points of each label to simulate their appearance at a long range.

\begin{figure}[tb]
    \centering
    
    \subfloat[\small Points' distribution resembles a Gaussian in the absence of the top face]{
        \includegraphics[width=0.42\linewidth]{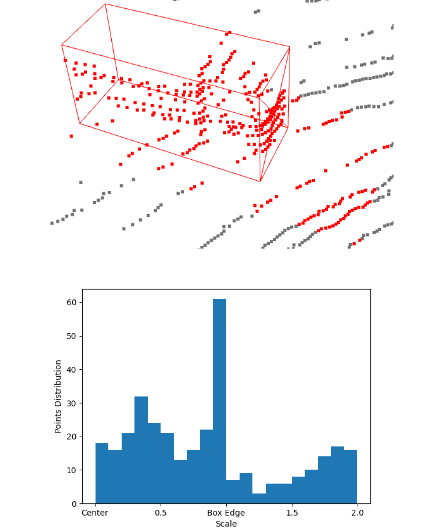} 
    }
    \hfill
    \subfloat[\small Points' distribution is uniform in the presence of the top face]{
        \includegraphics[width=0.42\linewidth]{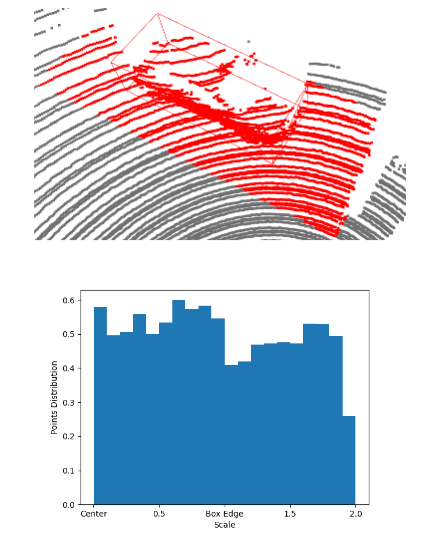} 
    }
    \caption{Distribution of points in the proximity of two detections with and without the top face, measured by \textit{scale}, calculated according to \cite{luo2023reward}.}
\label{fig:case_against_drift}
\end{figure}

The rate dropping procedure of OYSTER is tailored to LiDAR mounted on the roof of vehicles, making it unsuitable to apply to RSUs.
In contrast to prior works, we simultaneously resolve the restriction to dynamic objects and discovery range thanks to the aggregation of point clouds from multiple RSUs and multiple time-steps, offering dense and complete point clouds of the scene.
We further develop a refinement module based on multi-object tracking and point clouds registration to obtain better estimation of objects' dimension and poses.
Compared to the refinement of OYSTER which is also based on tracking, ours goes one step further to correct their poses by solving a least-square optimization.
DRIFT \cite{luo2023reward} extends MODEST by improving its performance through a reward ranked fine-tuning.
Particularly, DRIFT uses a reward function based on several heuristics for ranking an unsupervised model's detections and keeps only high ranked ones to use as labels during self-training.
Its most important heuristic - \textit{alignment reward}, which gives high scores to detections having points scattering near their side faces, is not suitable for RSUs' point clouds where objects' top faces are visible.
As illustrated in Fig.\ref{fig:case_against_drift}, points on objects' top face render the assumption behind the alignment reward that good detections have points distributed according to the Gaussian distribution invalid.
Our method, taking a different approach, performs fine-tuning directly on manually-generated labels.

\subsection{Semi-Supervised Object Detection}
In parallel to our work, semi-supervised learning (SSL) \cite{zhao2020sess, wang20213dioumatch, yin2022semi, liu2023hierarchical} also addresses label-efficient 3D object detection. 
Assuming a part of the training set is labeled, these methods train a teacher model in a fully supervised manner, then use it to generate labels, referred to as \textit{pseudo labels}, for the unlabelled data which are used for training the student model.
These methods focus on improving the quality of pseudo labels 
Wang et al~\cite{wang20213dioumatch} address this by filtering low confidence and poorly localized pseudo labels using detection confidence scores and estimated Intersection-over-Union, respectively.
\cite{yin2022semi} targets the recall rate of the teacher which they improve using test-time augmentation and an ensemble of models made of the teacher's weights obtained at difficulty epochs.
\cite{liu2023hierarchical} uses a dual-threshold strategy to filter low quality pseudo labels of difficulty classes.
As our method is based on unsupervised object discovery, the availability of labeled data is optional.

MS3D \cite{tsai2023ms3d} and its follow-up work \cite{tsai2023ms3d++} take a different setting where supervision comes as manually-generated labels of different domains (i.e., datasets).
They use an ensemble of models trained on manually-labeled data, referred to as experts, to generate pseudo labels.
Pseudo labels are then refined by tracking to exclude dynamic objects and obtain a better dimension estimation of static objects. 
While being similar in using tracking for discovered objects refinement, the role of tracking in our method is not to exclude dynamic objects but to jointly improve dimension and pose estimation.

%%%%%%%%%%%%%%%%%%%%%%%%%%%%%%%%%%%%%%%%%%%%%%%%%%%%%%%%%%%%%%%%%%%%%%%%%%%%%%%%
\section{Method}
\subsection{Problem Definition}
We target the scenario where an intersection is covered by at least one RSU, each of which has at least one LiDAR.
These RSUs are infrastructure; therefore, we assume they are well localized in a common frame of reference and obtain point clouds in synchronization.
The input to our object discovery is the aggregation of point clouds from each of the RSUs, which is possible because the object discovery takes place offline using recorded data.
A small part of the training set is manually annotated for hyperparameters tunning and model fine-tuning.
The validation set, on the other hand, is manually annotated entirely.

\subsection{Notations}
A point cloud is a set of 3D points, denoted by $\mathcal{P} = \{\mathbf{p}_j=[x, y, z]\}$.
A cluster, $\mathcal{C}$, is a subset of the point cloud $\mathcal{P}$ containing points from a single object.
An object is localized by its bounding box $\mathbf{b} = \left[c_x, c_y, c_z, w, l, h, \theta, v_x, v_y \right]$ 
which is parameterized by the 3D coordinate of its center $[c_x, c_y, c_z]$, its dimension $[w, l, h]$, its yaw angle $\theta$, and its velocity on the horizontal plane $[v_x, v_y]$.
In the following, we use \textit{object} to refer to both the cluster $\mathcal{C}$ of the foreground object and the bounding box $\mathbf{b}$ that encapsulates this cluster.

\begin{figure*}[tb]
    \centering
    \includegraphics[width=0.95\linewidth]{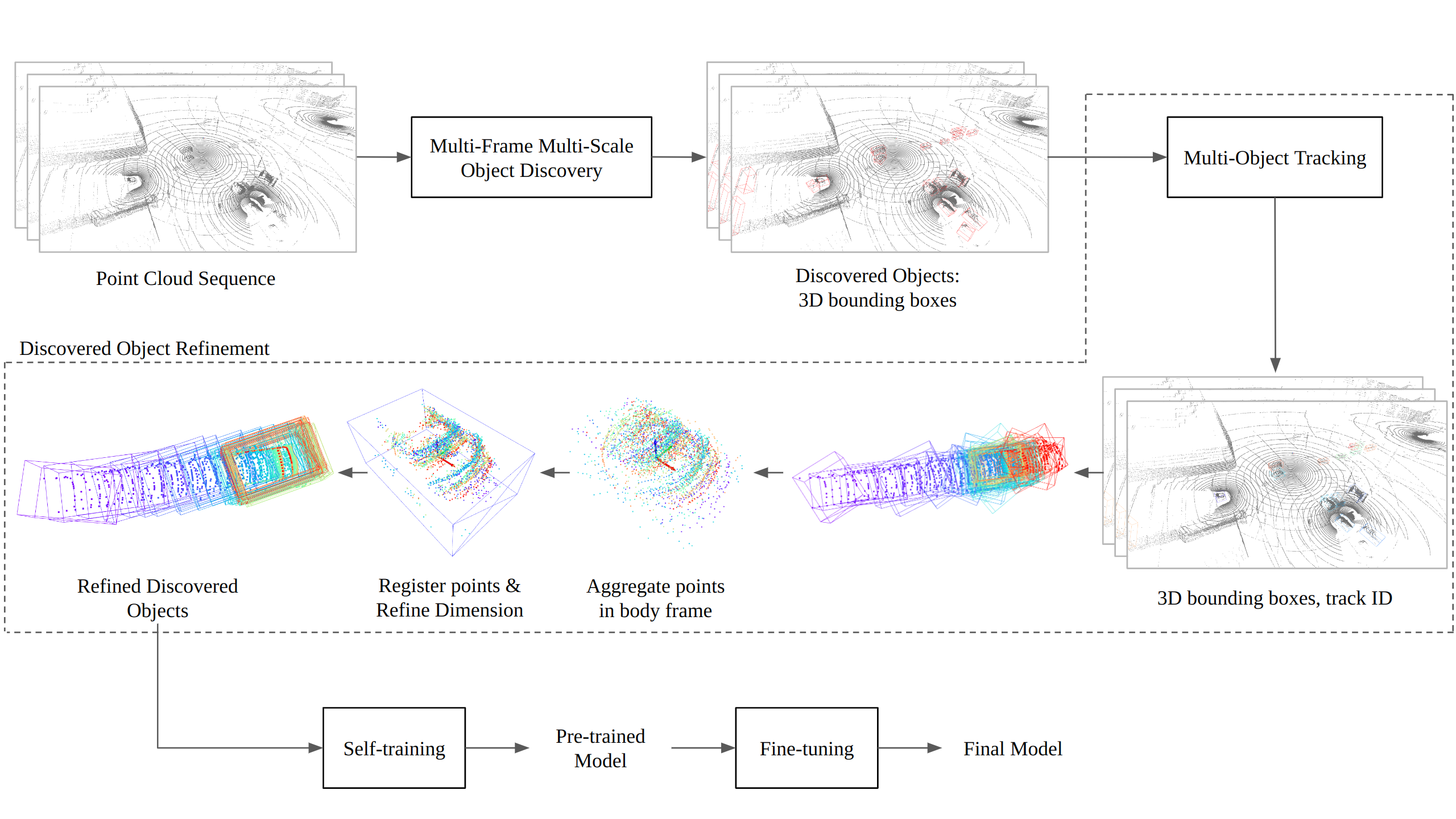}
    \caption{Overview of our method}
\label{fig:auto_rsu_method}
\end{figure*}

%%%%%%%%%%%%%%%%
Our method, illustrated in Fig.~\ref{fig:auto_rsu_method}, comprises four stages: multi-frame, multi-scale object discovery, refinement, self-training, and fine-tuning.
We will present the details of each module in the following sub-sections.

%%%%%%%%%%%%%%%%
\subsection{Object Discovery}
\begin{figure}[tb]
    \centering
    
    \subfloat[Object discovery results using one point cloud. Points are colored according to their cluster index. Black points are outliers.]{
        \includegraphics[width=0.96\linewidth]{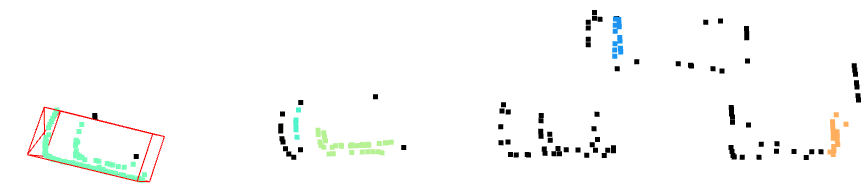} 
        \label{fig:det_by_cluster_1pc}
    }\\ 
    \vfill
    
    \subfloat[Object discovery results using the aggregation of points at \textcolor{blue}{(t-1)}, \textcolor{green}{t}, and \textcolor{red}{(t+1)}.]{
        \includegraphics[width=0.96\linewidth]{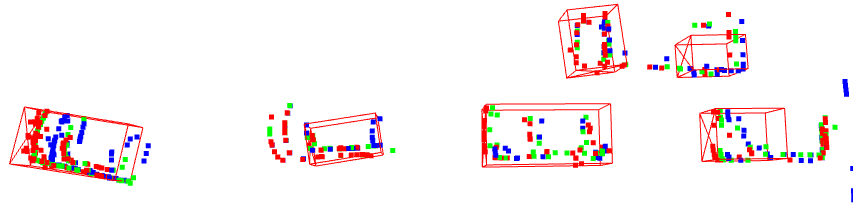} 
        \label{fig:det_by_cluster_3pcs}
    }

    \caption{Comparison of discovered objects using one point cloud, and the concatenation of three point clouds.}
\label{fig:det_by_cluster_multiframe}
\end{figure}

The typical object discovery pipeline consists of four sequential steps: ground removal, cluster detection, bounding box fitting, and filtering based on cluster dimensions \cite{you2022learning}.
The second step using density-based clustering algorithms such as DBSCAN result in sensitivity of this pipeline to point density, which is illustrated in Fig.~\ref{fig:det_by_cluster_multiframe}.

We ensure a high point density by aggregating point clouds spatially from multiple RSUs and temporally from multiple timesteps.
While the spatial aggregation is rather straightforward thanks to the static positions of RSUs, the temporal aggregation requires the treatment of noise in terms of tails from dynamic objects, which is caused by the change of objects' locations from one timestep to another.
We resolve this issue by aligning point clouds using scene flow.

\subsubsection{Scene Flow Estimation}
% summarize of ICP-flow
We use the unsupervised method ICP-flow \cite{lin2024icp} for scene flow estimation.
The method first detects clusters using HDBSCAN \cite{campello2013density} in two ego-motion-compensated point clouds, $\mathcal{P}^t$ and $\mathcal{P}^{t'}$, then matches clusters found in $\mathcal{P}^t$ to those in $\mathcal{P}^{t'}$ by solving a linear assignment problem representing a cost matrix $\mathbf{M}$.
An element at row $i$ and column $j$ of the cost matrix $\mathbf{M}$ represents the inlier ratio of registering cluster $\mathcal{C}_i^t$ found in $\mathcal{P}^t$ to cluster $\mathcal{C}_j^{t'}$ found in $\mathcal{P}^{t'}$ using ICP \cite{besl1992method}.
The transformation resulting from the registration of a pair of matched clusters is used to displace points in cluster $\mathcal{C}_i^t$ so that their scene flows are calculated as their displacement.

\subsubsection{Object Discovery At Multiple Scales}
We observe that a large vehicle (e.g., bus, truck) can exhibit large gaps between points due to the large object size and sparse lidar patterns, especially when it is at a long distance from the RSU.
These gaps cause DBSCAN to segment its points into multiple small clusters, which would be then disregarded because of their small sizes.
Our solution to this issue is to sequentially scaling the input point cloud with different factor from large to small and clustering on each scale to discover regular to large vehicles.
The motivation is that clustering at a large scale picks up clusters of regular-sized vehicles.
On the other hand, scaling the input point cloud with a small factor makes large groups of points smaller, thus easier to detect clusters of large vehicles.
A failure mode of our scaling approach is when two large vehicles are near others.
In this case, both of them are not detected on the large scales and are grouped together by DBSCAN on a small scale, resulting in an abnormally large vehicle.
This large detection is removed by the dimension-based filtering step, thus causing two false negatives.
Our multi-frame multi-scale object discovery is presented in Alg.~\ref{alg:multi_scale_cluster}.

\begin{algorithm}[htb]
    \caption{Object Discovery}
    \label{alg:multi_scale_cluster}
    \DontPrintSemicolon
      
    \KwInput{ \\
        $\mathcal{P}^t$:  the point cloud at the current timestep $t$ \\
        $\left\{ \mathcal{P}^{t_i} \vert i = 1,..., k \right\}$: $k$ point clouds at different timesteps
    }
    \KwOutput{
        $\mathcal{B} = \left\{\mathbf{b}_i\right\}$: discovered objects in $\mathcal{P}^t$
    }

    \BlankLine

    \For{
        $i = 1,...,k$
    }{
        $\mathcal{S}^i =$ scene\_flow$\left( \mathcal{P}^{t_i}, \mathcal{P}^t \right)$  
        
        \tcp{with $\mathcal{S}^i:=\left\{ \mathbf{f}_j = [f_x, f_y, f_z] \right\}$}

        $\mathcal{P}^{t_i} =$ translate$\left( \mathcal{P}^{t_i}, \mathcal{S}^i \right)$

        $\mathcal{P}^t = \mathcal{P}^t \cup \mathcal{P}^{t_i}$
    }

    $\mathcal{P}^t =$ remove\_ground($\mathcal{P}^t$)

    $\mathcal{B} = \emptyset$

    \For{$s$ in list of scales
    }{
        $\mathcal{P}_s =$  scale$\left( \mathcal{P}^t, s \right)$

        $\left\{ \mathcal{C}_i \right\} =$ clustering$(\mathcal{P}_s)$

        \For{
            each cluster $\mathcal{C}_i$
        }{
            $\mathbf{b}_i =$ box\_fitting$\left[\text{inverse\_scale}\left( \mathcal{C}_i, s \right)\right]$

            $\mathcal{P}^t = \mathcal{P}^t \setminus \text{inverse\_scale}\left( \mathcal{C}_i, s \right)$

            $\mathcal{B} = \mathcal{B} \cup \mathbf{b}_i$
        }
    }
\end{algorithm}

%%%%%%%%%%%%%%%%
\subsection{Objects Refinement}
The cluster-driven nature of our object discovery method makes detection inherently challenging when objects are only partially visible. 
% as illustrated in Fig.~\ref{fig:tracklet_varying_size}.
% \begin{figure}[tb]
%     \centering
%     \includegraphics[width=0.96\linewidth]{img/tracklet_varying_size.png}
%     \caption{Bounding boxes of a tracklet have varying sizes due to the reduced number of points. Points are color-coded according to their timestamps.}
%     \label{fig:tracklet_varying_size}
% \end{figure}
We leverage the intuition that an objects' visibility can improve over time due to their motion, we devise a refinement module based on an objects' trajectory, which we refer to as a \textit{tracklet}.
Formally, a tracklet $\mathcal{T}$ is a sequence of bounding boxes, refered to as \textit{instances}, representing the pose and dimension of an object at different timesteps when this object is observable.
\begin{equation}
    \mathcal{T} = \left\{
        \mathbf{b}^{t_i} = \left[c_x, c_y, c_z, l, w, h, \theta, v_x, v_y\right] \vert i = 1,...,k
    \right\}
\end{equation}
Once a tracklet is formed, the dimension of the corresponding object and the refined pose of its instances are calculated following the procedure in Sec.~\ref{sec:disco_refinement_new_dim} and Sec.~\ref{sec:disco_refinement_updat_pose}.

%%%%
\subsubsection{Multi-Object Tracking}
To form tracklets, we use a track-by-detection algorithm \cite{Weng2020_AB3DMOT}.
This algorithm extends a tracklet, formed at timestep $(t-1)$, to the current timestep $t$ by finding the bounding box $\mathbf{b}_t$ that matches with the prediction of its state.
The prediction of a tracklet's states is done using a constant velocity model.
The matched bounding box is used to update the prediction of the tracklet's state according to the formulation of the Kalman filter.

\subsubsection{Refining Dimension} \label{sec:disco_refinement_new_dim}
We build a complete reconstruction of an object from its tracklet by first aggregating points $\left\{ ^{w}\mathbf{p}^{t_i}_j \right\}$ residing inside the bounding box of instance $\mathbf{b}^{t_i}$, then transforming aggregated points from the world frame $\mathcal{F}_w$ to the object's body frame $\mathcal{F}_o$, and finally aligning them using ICP.
\begin{equation}
    \left\{ ^{o}\mathbf{p}^{t_i}_j \right\} = \text{ICP}\left( 
        \text{rigid\_body\_transform}\left(
            ^{w}\mathbf{T}_{o}^{-1}, \left\{ ^{w}\mathbf{p}^{t_i}_j \right\}
        \right)  
    \right)
    \label{eq:instance_point_in_body}
\end{equation}
here, $^{w}\mathbf{T}_{o}$ is the rigid body transformation that transforms points from the body frame $\mathcal{F}_o$ of the instance $\mathbf{b}^{t_i}$ to the world frame $\mathcal{F}_w$ 
\begin{equation}
    ^{w}\mathbf{T}_{o} = \begin{bmatrix}
        \cos \theta^{t_i} & -\sin \theta^{t_i} & 0 & c_x^{t_i} \\
        \sin \theta^{t_i} & \cos \theta^{t_i} & 0 & c_y^{t_i} \\
        0 & 0 & 1 & c_z^{t_i} \\
        0 & 0 & 0 & 1
    \end{bmatrix}
\end{equation}
A bounding box is fit to the resulting points to obtain the dimension of the object.

\subsubsection{Updating Object Pose} \label{sec:disco_refinement_updat_pose}
We "align" the pose (i.e., position and orientation) of instance $\mathbf{b}^{t_i}$ by seeking the transformation $^{w}\mathbf{T'}_o$ that best explains the coordinates of the instance's points in the object frame $\mathcal{F}_o$ as computed by Eq.~\eqref{eq:instance_point_in_body}.
We solve the following optimization problem

\begin{equation}
    ^{w}\mathbf{T'}_o = \operatorname*{argmin}_{^{w}\mathbf{T}_o} \sum_{i} \norm{
        ^{w}\mathbf{T}^{-1}_o \cdot ~ ^{w}\underline{\mathbf{p}_j^{t_i}} ~ - ~ ^{o}\underline{\mathbf{p}_j^{t_i}}
        }
    \label{eq:pose_refine}
\end{equation}
here, $\underline{\mathbf{p}}$ denotes the homogeneous coordinate of the 3D point $\mathbf{p}$.
The entire refinement pipeline is presented in Alg.~\ref{alg:disco_refine}.

\begin{algorithm}[htb]
    \caption{Object Refinement}
    \label{alg:disco_refine}
    \DontPrintSemicolon

    \KwInput{
        $\mathcal{T} = \left\{
            \left( \mathbf{b}^{t_i}, \left\{^{w}\mathbf{p}_j^{t_i}\right\}   \right) 
        \right\}$
        \tcp{instances of a tracklet and their points}
    }

    \BlankLine

    \For{
        each $t_i$
    }{
        $\left\{^{o}\mathbf{p}_j^{t_i}\right\} = \text{rigid\_body\_transform}\left(
            ^{w}\mathbf{T}_o, \left\{^{o}\mathbf{p}_j^{t_i}\right\}
        \right)$
    }

    $\left\{^{o}\mathbf{p}_j^{t_*}\right\}$ = points of the largest instance
    
    $\mathcal{P} = \left\{^{o}\mathbf{p}_j^{t_*}\right\}$
    
    \For{
        each $t_i \neq t_*$
    }{
        $\left\{^{o}\mathbf{p}_j^{t_i}\right\} = \text{ICP}\left(
            \left\{^{o}\mathbf{p}_j^{t_i}\right\}, \mathcal{P}
        \right)$  ~~\tcp{Eq.\eqref{eq:instance_point_in_body}}

        $\mathcal{P} = \mathcal{P} \cup \left\{^{o}\mathbf{p}_j^{t_i}\right\}$
    }

    \BlankLine
    
    $\mathbf{b}_* = \text{box\_fitting}\left( \mathcal{P} \right)$ ~~\tcp{refine dimension}

    \BlankLine

    \For{
        each $t_i$
    }{
        new pose  = solve Eq.~\eqref{eq:pose_refine}
    }

\end{algorithm}

%%%%%%%%%%%%%%%%
\subsection{Self-Training} \label{sec:self_training}
We use the discovered objects to start a self-training process, where a learning-based detection model iteratively improves its performance.
In the first iteration, a detection model $\mathfrak{D}^0$ is trained from scratch with the discovered objects as labels.
Once $\mathfrak{D}^0$ converges, the high-confidence detections from $\mathfrak{D}^0$ on point clouds from the training set act as labels in the second iteration, which is why they are referred to as \textit{pseudo labels}. 
This process is repeated until the maximum number of iterations is reached or the performance saturates.
 
% \begin{algorithm}[htb]
%     \caption{Self Training}
%     \label{alg:self_training}
%     \DontPrintSemicolon

%     \KwInput{

%         Point clouds of the training set $\{\mathcal{P}^{j}\}$
        
%          Discovered objects $\left\{ \mathcal{D}^{j} = \left\{ \mathbf{b}^j_i \right\}\right\}$
        
%         number of self-training rounds $r$
%     }

%     \KwOutput{A trained detector $\mathfrak{D}^r$}
    
%     \BlankLine

%     $\mathfrak{D}^0 =$ train$\left( \{\mathcal{P}^{j}\}, \{\mathcal{D}^{j}\} \right)$ 

%     \For{$i$ in \{1, 2, ..., r\}}{
%         detections = inference$\left(\mathfrak{D}^{i - 1}, \{\mathcal{P}^{j}\} \right)$

%         pseudo labels = get\_high\_confidence(detections)

%         $\mathfrak{D}^i =$ train$\left( \{\mathcal{P}^{j}\}, \text{pseudo labels}\right)$

%     }
% \end{algorithm}

%%%%%%%%%%%%%%%%
\subsection{Fine-Tuning} \label{sec:method_fine_tuning}

The benefit of self-training is that the model is able to learn patterns associated with the presence of objects and thus is able to detect objects that are missed in the discovery phase.
This extrapolation ability is at the cost of false positive detections.
At any self-training round, the detection filtering based on confidence score can not prevent these false positives from propagating to the next round because some of them have relatively high confidence. 
As a result, the model consolidates its false belief about the appearance of foreground objects.
The correction of this false belief requires human intervention which we carry out by fine-tuning the model trained in the self-training phase on the manually-labeled part of the training set.

Beside the straightforward implementation of fine-tuning that is to re-train the model from the pre-trained weights on new data, we introduce a scheme that mixs self-training and fine-tuning to obtain higher performance with the same amount of labeled data.
Particularly, given a self-trained model, we repeat $m$ times the following process (1) one-iteration self-training, and (2) fine-tuning.
Our scheme is motivated by the observation that self-training increases the model's recall rate, resulting in more true positives and false positives.
Fine-tuning helps remove those false positives, thus increasing the final precision.

%%%%%%%%%%%%%%%%

%%%%%%%%%%%%%%%%%%%%%%%%%%%%%%%%%%%%%%%%%%%%%%%%%%%%%%%%%%%%%%%%%%%%%%%%%%%%%%%%
\section{EXPERIMENTS}
\textbf{Datasets.} 
We validate our method on two datasets: V2X-Sim 2.0~\cite{li2022v2x} and A9-Intersection, or A9 for short, \cite{zimmer2023a9}.
V2X-Sim 2.0 is a synthetic dataset made with CARLA \cite{dosovitskiy2017carla}.
It contains 100 sequences, each of which has 100 samples containing point clouds obtained by 1 RSU positioned at the center of an intersection of a town of CARLA and up to 5 connected vehicles.
Due to the absence of a dataset containing multiple RSUs, we use the connected vehicles in V2X-Sim to simulate the scenario where an intersection is covered by more than 1 RSUs.
We put 8900 data samples collected at Town 04 and 05 to the training set and 1100 samples from Town 03 to the validation set.
A9 is a real-world dataset made of two RSUs positioned at 7 meters height in an intersection in Garching, Germany.
Each RSU has an Ouster OS1-64 LiDAR to obtain point clouds of the intersection at 10Hz and in synchronization with the other.
A9 has 1920 samples for training and 240 samples for validation.

\textbf{Evaluation metric.} 
As we design our approach to detect vehicles, we merge the ground truth of every wheeled-vehicle class to a unified \textit{vehicle} class.
We use mean Average Precision (mAP) and Detection Score (NDS) of the nuScenes dataset~\cite{caesar2020nuscenes} as the evaluation metrics because they correlate well with downstream driving tasks \cite{schreier2023offline}.
NDS is the weighted sum of mAP and positive metrics including translation error, scale error, orientation error, and velocity error, averaged over all true positive detections.
In the following tables, the best and second best performance in each metric are shown in \textbf{bold} and  \underline{underline}, respectively.

\textbf{Implementation.} 
We demonstrate our approach using VoxelNext~\cite{chen2023voxelnext}. 
Nevertheless, our findings are applicable to other detection models.
To facilitate reproducibility, our implementation is based on the open-source code from OpenPCDet~\cite{openpcdet2020}.
We use the multi-object tracking implemented by \cite{immortaltrack}.
The hyper-parameters of VoxelNext are kept identical to the default setting of OpenPCDet, with the exception of reducing the number of epochs to 10 and setting the total batch size to 12.
Following the evaluation protocol of nuScenes for wheeled-vehicle class, the detection range is set to 50 meters from the center of the intersection.
The number of self-training iterations is 3 on both datasets.
The training takes place on a single NVIDIA A6000 GPU.

%%%%%%%%%%%%%%%%%%%%%%%%%%%%%%%%%%%%%%%%%%%%%%%%%%%%%%%%%%%%%%%%%%%%%%%%%%%%%%%%%
\subsection{Object Discovery Results}
We compare our object discovery method against two baselines: DBSCAN and a state-of-the-art object discovery method - MODEST \cite{you2022learning}. 
Our implementation on V2X-Sim dataset uses 3 frames (i.e. point clouds); therefore, we also feed up to 3 frames to the two baselines.
As MODEST is originally developed to discover objects using multiple traversals of the same route, we adapt this method to the setting of RSUs by reasoning that each RSU's point cloud is equivalent to a traversal because it is stationary.
Similar to our method, the two baselines use the aggregation of point clouds from every available RSU as their input, thus having dense and more complete point clouds of the scene.

The comparison result in Tab.~\ref{tab:comparison_obj_disco} shows that our method outperforms the best setting of DBSCAN of MODEST by 122.7\% and 42.3\% in terms of mAP.
To highlight the ability of detecting more ground truth at high precision, we present the comparison of recall rate computed by matching discovered objects with ground truth at the Intersection-over-Union threshold 0.3.
It is worth noticing that the two baselines and our discovery method do not produce confidence score for discovered objects;, therefore, the calculation of recall rate in Tab.~\ref{tab:comparison_obj_disco} performs regardless of this quantity.
The underperformance of MODEST compared to ours can be explained by its restriction to dynamic objects.
As RSUs are positioned at intersections where only traffic in one direction is allowed to move, their environments contain less dynamic objects compared to those of vehicles which MODEST is designed for.
Therefore, the exclusion of static objects limit MODEST's recall rate. 

\begin{table}[htb]
    \caption{comparison of different object discovery methods on the validation set of V2X-Sim.
    }
    \label{tab:comparison_obj_disco}
\centering
\begin{tabular}{l c c c}
    \toprule
                         &    mAP               &   NDS             &   Recall      \\  
    \midrule
    DBSCAN - 1 Frame     &   12.46              &   27.03           &   29.98       \\
    DBSCAN - 2 Frames    &   14.30              &   40.28           &   38.47       \\
    DBSCAN - 3 Frames    &   16.70              &   36.94           &   44.34       \\
    \midrule
    MODEST - 2 traversals\footnotemark[1]       &   23.76              &  38.98            &   43.13       \\
    MODEST - 3 traversals\footnotemark[1]       & 26.14             &  \underline{40.41}            &   45.87       \\
    \midrule
    Our object discovery       &   35.82              &   35.12           &   \textbf{70.71}       \\
    + refinement               &   \textbf{37.19}              &   \textbf{51.88}           &   \underline{53.70}       \\
    
    \bottomrule
\end{tabular}
\end{table}

\footnotetext[1]{Our adaptation to the setting of RSU data}

The importance of the refinement module is indicated by 47.7\% improvement in NDS, from 35.12 to 51.88, when it is integrated into our object discovery module.
This improvement is due to a 43.2\%, 66.0\%, and 99.4\% reduction in translation error, scale error, and orientation error, respectively, averaged over all true positive detections.
The refinement module also results in a 24\% drop of recall rate, from 70.71 to 53.70. 
This is because of the filtering of tracklets having few instances, which are predominantly tracklets of spurious detections.
While reducing the number false positives, this filtering process inevitably removes some true positives, especially when the tracking algorithm fails to make the association between tracklets and detections. 

%%%%%%%%%%%%%%%%%%%%%%%%%%%%%%%%%%%%%%%%%%%%%%%%%%%%%%%%%%%%%%%%%%%%%%%%%%%%%%%%%
\subsection{Self-Training Results} \label{sec:self_training_results}  % TODO: continue refinement here
\begin{figure}[tb]
    \centering
    \subfloat[V2X-Sim]{
        \includegraphics[width=0.45\linewidth]{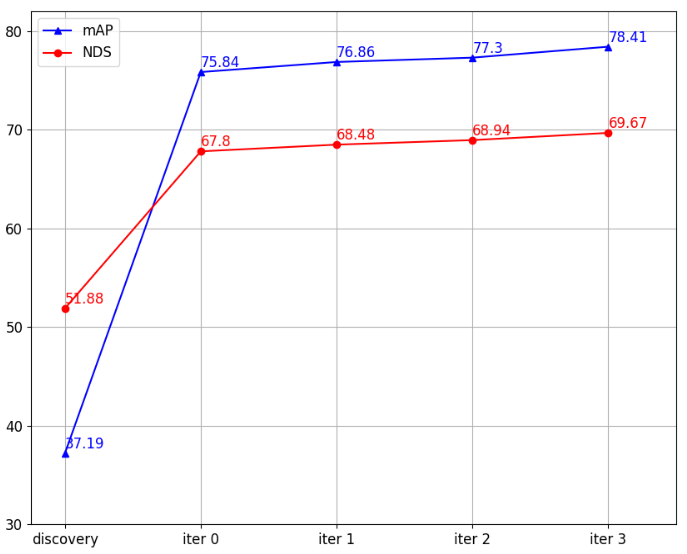} 
        \label{fig:self_training_v2x}
    }
    \subfloat[A9]{
        \includegraphics[width=0.45\linewidth]{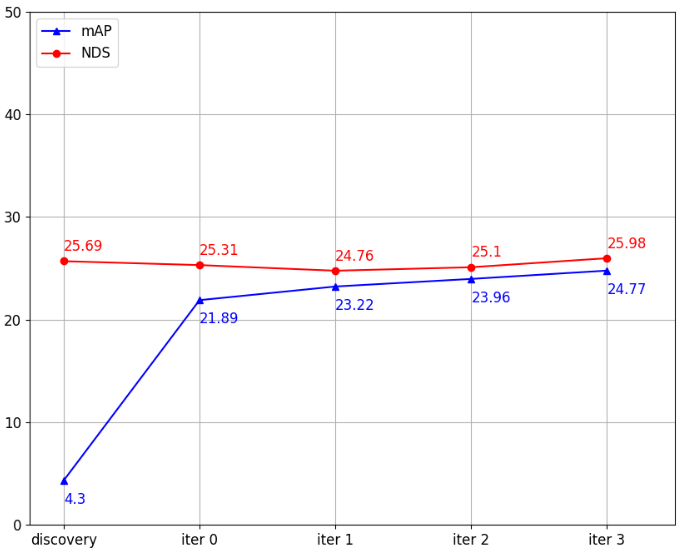}    
        \label{fig:self_training_a9}
    } 
    
    \caption{The evolution of detection performance during self-training.}
\label{fig:self_training}
\end{figure}

Fig.~\ref{fig:self_training} shows the evolution of detection performance during the self-training process.
The most substantial improvement, by 103.9\% (from 37.19 to 75.84) on V2X-Sim and 409.1\% (from 4.3 to 21.89) on A9 in terms of mAP, takes place at the first iteration where discovered objects are used as labels, showing the ability of capturing patterns in the training data of deep learning models.
As the self-training progresses, the rate of improvement quickly reduces to roughly 1 mAP per iteration, showing that the set of pseudo labels is consistent.

Unlike mAP, NDS shows a lower rate of improvement during our experiments in V2X-Sim dataset while being stagnated in A9 dataset.
The breakdown of NDS indicates that all positive metrics remain unchanged throughout the self-training process in V2X-Sim dataset.
Therefore, its improvement is solely due to the increase of mAP.
The evolution of positive metrics in A9 dataset share the same pattern except for the orientation error.
This error increases from 1 degree of discovered objects to 60 degrees in the first self-training iteration, then remains unchanged during subsequent iterations.
This shows that the self-trained model, while successfully detecting the pattern that indicating the presence of objects in point clouds, encounters difficulty in learning to precisely predict their location, dimension and orientation,.
This observation highlights the importance of precisely localizing objects in the discovery phase, which we achieve with the refinement module.

The comparison of our method against MS3D++~\cite{tsai2023ms3d++} on the V2X-Sim dataset is shown in Tab.~\ref{tab:comparison_ms3d}.
Here, \cite{tsai2023ms3d++} uses an ensemble of two VoxelRCNN \cite{deng2021voxel} trained fully supervised on the entire training set of nuScenes \cite{caesar2020nuscenes} and Lyft dataset, respectively.
As the input to our object discovery consists of three frames, we also pass three frames to \cite{tsai2023ms3d++} for generating pseudo labels.
Benifiting from large amount of manully-generated labels in nuScenes and Lyft, \cite{tsai2023ms3d++} achieves almost double the precision of ours in the object discovery phase.
However, we surpass their performance after the self-training phase due to the domain gap between RSUs'data and autonomous driving data, caused by the different appearance of objects on RSUs' point clouds due to their elevated height and multi views.

\begin{table}[htb]
    \caption{comparison between our self-trained model and MS3D on the validation set of V2X-Sim.}
    \label{tab:comparison_ms3d}
\centering
\begin{tabular}{l c c}
    \toprule
                            &    mAP                &   NDS           \\  
    \midrule
    \cite{tsai2023ms3d++} - pseudo labels    &   68.41               &   79.34           \\
    \cite{tsai2023ms3d++} at iter 3          &   71.20               &   72.00           \\
    \midrule
    Ours - discovered objects &  37.19               &   51.88           \\
    Ours at iter 3           &  78.41               &   69.67           \\
    \bottomrule
\end{tabular}
\end{table}

%%%%%%%%%%%%%%%%%%%%%%%%%%%%%%%%%%%%%%%%%%%%%%%%%%%%%%%%%%%%%%%%%%%%%%%%%%%%%%%%%
\subsection{Fine-Tuning Results}
\begin{figure}[tb]
    \centering
    \subfloat[Visual comparison between a correctly detected truck (left) and falsely detected bus stop (right)]{
        \includegraphics[width=0.96\linewidth]{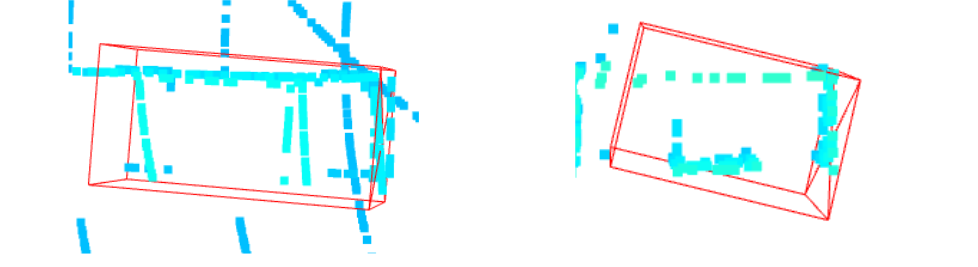} 
        \label{fig:sub_false_positive_wall_busstop_busstop}
    }\\ 
    \vfill
    \subfloat[Visual comparison between a correctly detected car (left) and falsely detected wall (right)]{
        \includegraphics[width=0.96\linewidth]{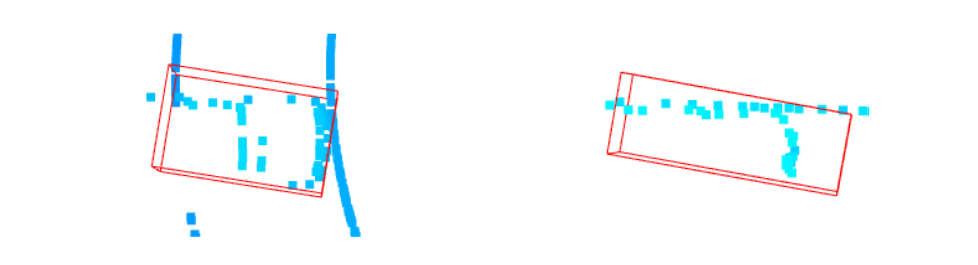}    
        \label{fig:sub_false_positive_wall_busstop_wall}
    } 
    
    \caption{False positive predictions of the self-trained model due to the similar appearance of vehicles and background objects.}
\label{fig:false_positive_wall_busstop}
\end{figure}

Sec.~\ref{sec:self_training_results} shows that self-training drastically improves the detection performance of the object discovery; however, the final self-trained model still underperforms the \textit{fully supervised model}.
Here, the the term "fully supervised model" refers to the model that is trained from randomly initialized weights on 100\% of the manually labeled training set as the fully supervised model.
This underperformance is due to (1) false positive detections caused by the appearance similarity between vehicles and background objects (e.g., wall and bus stops) illustrated in Fig.~\ref{fig:false_positive_wall_busstop} and (2) the difficulty in learning to predict precisely objects' location, dimension, and orientation as discussed in Sec.~\ref{sec:self_training_results}.

Tab.~\ref{tab:perf_finetune_scratch} shows that fine-tuning the self-trained model can overcome these two issues as an increase in the volume of manually labeled data available for fine-tuning leads to an increase in the model's performance.
With 20\% and 50\% of manually labeled data of the training set of V2X-Sim and A9, the fine-tuned model can reach the performance of the fully supervised one. 
Moreover, the comparison between fine-tuning the self-trained model and the one trained from randomly initialized weights given the same amount of manually labeled data shows a favorable result to fine-tuning, emphasizing the importance of object discovery and self-training especially when labeled data is scarce.

\begin{table}[htb]
    \caption{comparison of NDS between fine-tuning and training from randomly initialized weights}
    \label{tab:perf_finetune_scratch}
    \centering
    \resizebox{0.99\columnwidth}{!}{
\begin{tabular}{l  c c    c c}
\toprule
    Labeled data    & \multicolumn{2}{c}{V2X-Sim}                                             & \multicolumn{2}{c}{A9} \\
                    \cmidrule(lr){2-3}                                                        \cmidrule(lr){4-5}
    (\% of          & \multirow{2}{*}{Fine-tune} & Train from                                 & \multirow{2}{*}{Fine-tune} & Train from \\
    training set)   &                            & random init.                               &                            & random init. \\
    \midrule
    0              & 69.67   & 0.00                        & 25.98 & 0.00   \\
    1              & 78.51   & 36.54                       & 36.83 & 0.00   \\
    2              & 80.79   & 57.79                       & 46.46 & 0.00   \\
    5              & 85.14   & 77.77                       & 53.98 & 7.75   \\
    10             & 88.32   & 84.77                       & 64.04 & 15.75  \\
    20             & 89.33   & 89.75                       & 69.46 & 58.58  \\
    50             & \underline{91.03}   & 90.33                       & \underline{72.14} & 71.05 \\
    100            & \textbf{93.42}   & 91.00                       & \textbf{75.68} & 71.40 \\

\bottomrule
\end{tabular}
}
\end{table}

The comparison between the traditional fine-tuning and our scheme of mixing self-training and fine-tuning is presented in Tab.~\ref{tab:fine_vs_mixed_finetune}.
In this experiment, the amount of lableled data for both method is 100 point clouds which accounts for 1\% and 5\% of the training set of V2X-Sim and A9, respectively.
The number of times that self-training followed by fine-tuning is repeated, $m$, is 2.
Our mixed fine-tuning achieves higher performance than traditional fine-tuning given the same amount of labeled data and yields 99\% and 96\% performance of the fully supervised baseline, further proving the data-efficiency of our method.

\begin{table}
    \caption{comparison between the traditional fine-tuning and mixed fine-tuning \& self-training.
    }
    \label{tab:fine_vs_mixed_finetune}
    \centering
    \begin{tabular}{l l c c c}
    \toprule
    \multirow{2}{*}{Dataset}                  & \multirow{2}{*}{Method}               & Labeled data   & \multirow{2}{*}{mAP}       & \multirow{2}{*}{NDS} \\
                                            &                                         & (\% training set) & & \\
    \midrule
    \multirow{3}{*}{V2X-Sim} & Fine-tuning          & 1\%           & 87.71     &   78.51   \\
                             & Mixed self-training    & \multirow{2}{*}{1\%}           & \multirow{2}{*}{91.66}     &   \multirow{2}{*}{83.48}   \\
                             & \& fine-tuning       &               &           &           \\
                             & Fully supervised     & 100\%         & 93.00     &   91.00   \\
    \midrule
    \multirow{3}{*}{A9}      & Fine-tuning          & 5\%           & 54.59     &   53.99   \\
    & Mixed self-training    & \multirow{2}{*}{5\%}           & \multirow{2}{*}{67.10}     &   \multirow{2}{*}{67.81}   \\
                             & \& fine-tuning       &               &           &           \\
                            %  & Mixed fine-tuning    & 5\%           & 67.10     &   67.81   \\
                             & Fully supervised     & 100\%         & 69.64     &   71.40   \\
    
    \bottomrule
    \end{tabular}
\end{table}

%%%%%%%%%%%%%%%%%%%%%%%%%%%%%%%%%%%%%%%%%%%%%%%%%%%%%%%%%%%%%%%%%%%%%%%%%%%%%%%%%
\subsection{Real-World Experiments}
\begin{figure*}[htb]
    \centering
    \subfloat[]{
        \includegraphics[width=0.32\linewidth]{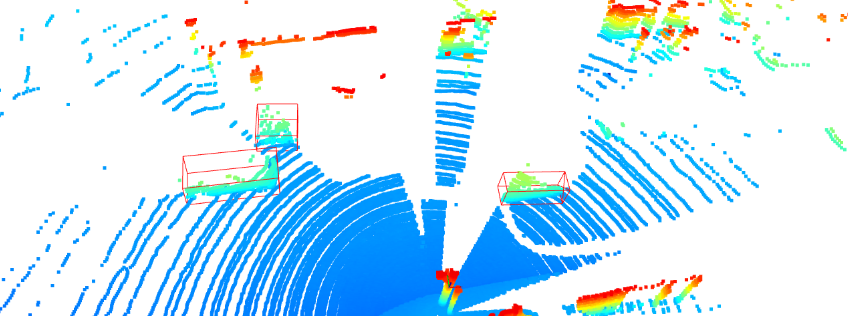} 
    } 
    \subfloat[]{
        \includegraphics[width=0.32\linewidth]{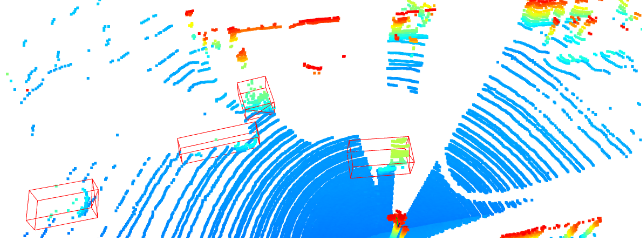}    
    } 
    \\
    \subfloat[]{
        \includegraphics[width=0.32\linewidth]{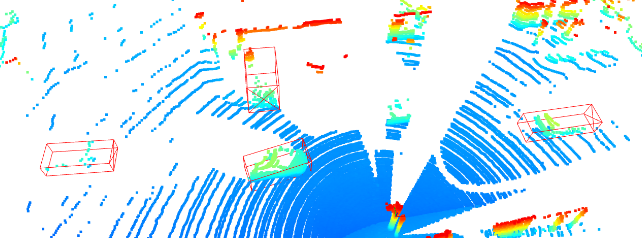} 
    } 
    \subfloat[]{
        \includegraphics[width=0.32\linewidth]{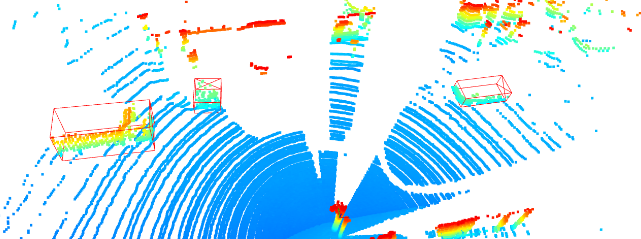}    
    }
    \caption{Visualization of discovered objects in our real-world data}
    \label{fig:sydney_data}
\end{figure*}

The dataset for real-world testing was collected using a RSU located at an urban intersection in Sydney, Australia. 
It has two LiDAR sensors: an Ouster OS1 64 beams and an Ouster Dome. 
The LiDARs are synchronized and phase-locked. 
The data for this qualitative evaluation was obtained during standard operation in moderate traffic conditions. 
Since there is no pre-existing annotated dataset for this specific setup, this paper presents these findings as qualitative results in Fig.\ref{fig:sydney_data}, showcasing the practical applicability of our method with real-world data.
An extended visualization of our object discovery and refinement module can be found in our video demonstration.

\section{Conclusion}
We propose the first autolabeling framework for RSUs' point clouds based on unsupervised object discovery.
In comparison to existing works, our framework comprises two novel modules: one for discovering objects at multiple scales through spatial and temporal aggregation of point clouds, and another for refining the discovered objects.
With only 100 manually-labeled point clouds provided for fine-tuning, the model pretrained with labels generated by our framework achieves a performance comparable to that of the model trained fully supervised on thousands of manually-labeled point clouds.
This results in a label-efficient object detection method for RSUs.

%%%%%%%%%%%%%%%%%%%%%%%%%%%%%%%%%%%%%%%%%%%%%%%%%%%%%%%%%%%%%%%%%%%%%%%%%%%%%%%%

\bibliographystyle{IEEEtran}
% \bibliography{biblio}

\end{document}